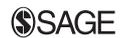

# Path planning for unmanned surface vehicle based on predictive artificial potential field

Jia Song[1], Ce Hao[1,2]  and Jiangcheng Su[1]

## Abstract
The path planning for high-speed unmanned surface vehicle raises more complicated requirements to reduce sailing time and save energy. In this article, a new predictive artificial potential field is proposed using time information and predictive potential to plan a smoother path. The principle of artificial potential field with vehicle dynamics and reachability in local minimum is studied. According to global and local path planning, the most state-of-the-art traditional artificial potential field and its drawback are analysed at first. Then we proposed predictive artificial potential field with three modifications: angle limit, velocity adjustment and predictive potential to improve the feasibility and flatness of the generated path. In addition, we compare the performance between traditional artificial potential field and predictive artificial potential field, where predictive artificial potential field successfully restricts the maximum turning angle, cuts short sailing time and intelligently avoids obstacle. From the simulation results, we also verify that predictive artificial potential field can solve concave local minimum problem and enhance the reachability in special scenario. Therefore, the more reasonable path generated by predictive artificial potential field reduces sailing time and helps conserve more energy for unmanned surface vehicle.

## Introduction

During the past decades, a growing number of unmanned surface vehicles (USVs) have been developed for economic, environmental and robotic research applications.[1,2] We, human, start to rely more on USVs to explore complicated and changeable maritime space. Because many laborious tasks like commercial shipping,[3] hydrographic survey[4] and ocean rescue[5] are more suitable for unmanned vehicle, especially for economic purpose, the USVs can replace human to execute repetitive tasks, which definitely save much cost and energy.[6]

Consequently, the importance of unmanned vehicles in the ocean exploration has been addressed by a large number of countries. For example, the wide applications of USV driving a rising demand vary from the scenario in North America, and the European market is also showing important development.[7] Besides, Google statistics show a strong increase in search volume for USVs.[8] The powerful

[1] School of Astronautics, Beihang University, Beijing, China
[2] Department of Mechanical Engineering, University of California, Berkeley, CA, USA

**Corresponding author:**
Ce Hao, School of Astronautics, Beihang University, 37 Xueyuan Road, Haidian District, Beijing, China.
Email: haoce@buaa.edu.cn

presence of Asia-Pacific consumers on Geo-matching could imply that many of this region's consumers are studying unmanned ground car technology, making it a powerful indicator of increasing market share in 2020.

However, with the development of autonomy technology, more requirements are raised by advanced USV. Undoubtedly, high speed is the primal index to evaluate the performance, while it may also introduce new difficulties like reliable path planning, dynamic constraints and energy saving.[9] The higher speed definitely improves penetrability, work range and efficiency, which totally enhances the utility of USV. Therefore, a secure and feasible path is necessary to serve high-speed USV.

Path planning is an overall task that composes global planning and local planning. The global planning is to navigate the USV sailing across multiple obstacles and reach the final target, which cares less about trajectory detail, while local planning concentrates more on collision avoidance and achieve a temporary goal.[10] Actually, the high speed influences more on the local planning as separated reef and tiny island are not totally marked on the map or unable to be detected within short distance.[11] Hence, in this article, we debate mainly on local path planning, namely collision avoidance to improve the velocity under the constraint of dynamic.

There are four basic path planning algorithms including grid-based method, line-of-sight (LOS) method, iterative method and artificial potential field (APF). The grid-based method includes A*,[12,13] Dijkstra[14] and fast marching[15] algorithm which search the shorted distance from the start point to the target. Methodically speaking, grid-based method searches iteratively around the whole map and consumes much computation. A precise map may improve accuracy but increases computing time exponentially. So, the grid-based method is more suitable for overall map planning with rough grid rather than exact local planning. Besides, LOS[16] method is achieved by geometric projection for minimization of the cross-track error to the path. But the LOS needs a former generated path, which makes it less efficient in online planning. In addition, the iterative algorithm composes a large family of optimization like genetic algorithm,[17,18] particle swarm optimization[19] and ant colony optimization.[20] The iterative algorithm basically utilizes stochastic optimization to explore the whole map. With strict constraints, the algorithm finally generates a perfect path after thousands of iterations. Therefore, only the offline planning for static obstacles would be effective by iterative algorithm. Finally, APF[21] is very popular with both global and local planning for its adjustable potential field. The APF navigates the USV to the final goal by continuous attractive and repulsive field; however, the local minimum might trap the USV and cause unreachability. To sum up, the first three methods do not cater to fast-moving USV for high computation, while the APF is preferred to generate trajectory with customized improvements.

Another essential concern for path planning is the authenticity of vehicle dynamic. So far, all navigation algorithms are not tightly integrated with the control strategy. For example, the grid-based algorithm may find a possible path with high-frequency vibration or a sharp turning, which is not corresponding with the dynamic model. Thus, we believe that trajectory variables like planar and angular velocity should be considered as well. Fortunately, the APF navigates continuously and updates in real time. It is suitable to modify APF with vehicle dynamic constraints.

However, whatever the algorithm is, it does not take into account the predictive ability. The main difference between human and intelligent algorithm is the perception and prediction to detour the obstacle in advance. We, human, are able to see where the obstacles are and take action long before coming close to the forbidden area, while the computer-based algorithms mechanically explore every possible path and avoid until the edge of collision. Undoubtedly, predictive ability helps the USV to react to a possible collision in future intelligently and actively.

In this article, we propose predictive artificial potential field (PAPF) method for USV path planning according to dynamic constraints and predictive ability. First, we adopt two modifications on turning angle limit (AL) and velocity adjustment (VA), which improve the smoothness of the path and cut short sailing time. Besides, the predictive potential field improvement is proposed to compensate traditional potential field and avoid obstacle in advance. Finally, we combine the modifications as PAPF, which definitely generates path with small turning angle, higher speed and smoother track. The simulation shows that the PAPF can solve the concave local minimum problem and enhance the reachability.

The structure of this article is as follows. Section 'Related works' introduces the relative research and development of USV and APF. Section 'Method and result' analyses the traditional APF and proposes the structure of PAPF and its navigation on USV. Then, the section 'Discussion' discusses the principle and the possible extension of PAPF for USV obstacle avoidance. Finally, the section "Conclusion" summaries the content of the whole article.

## Related works

The first APF was proposed by Khatib[22] in 1986 for global path planning. Later, Warren[21] expanded APF to multi-obstacles and adjusted the shape of repulsive field. However, there is a major problem in the local path planning using the APF approaches, which is that the local minimum can trap an USV before reaching its goal. The local minimum problem is sometimes inevitable in the local path planning because the USV only can detect local information of obstacles. In other words, the robot cannot predict local minimum before experiencing the environments.

To solve the local minimum problem, Uyanik et al.[23] adopted negative potential function with many complex configurations to set. Besides, Guo et al.[24] combined the improved APF with navigation function, while it is an indirect method with plenty of artificial adjustments. Furthermore, Song et al.[25] utilized biogeography-based optimization to pass through a very thin tunnel between obstacles, which usually should be regarded as a large obstacle to ensure security. In addition, Liu et al.[26] utilized configuration-oriented APF for three-dimensional avoidance. More interestingly, Wu et al.[20] employed APF to accelerate ant colony optimization, but the drawbacks of both grid-based and iterative algorithms exposed.

More recently, Chen et al.[27] applied optimal control theory into APF while several sharp turnings still remained. Progressively, Lyu and Yin[28] proposed the predictive idea for obstacle avoidance, which mostly inspired us to adopt predictive potential. But unfortunately, the predictive idea was not widespread as an independent approach. Besides, Manzini[29] in his thesis presented the piecewise attractive field to reduce the square increasement, which is adopted as the traditional APF in the following section. Another inspiration of angle-based predictive potential came from Song et al.[30] In this article, angular distance was considered as a variable that limits the movement of USV.

With the development of APF, more and more people started to incorporate dynamic model and control strategy and adopt the predictive idea to improve performance. Therefore, we are inspired to utilize APF with velocity constraints and predictive potential field to generate a smoother and reasonable path, which helps save energy and extends cruising range.

## Method and result

### Traditional artificial potential field

The traditional artificial potential field (TAPF) has developed over decades with different modifications. According to the special marine condition like island navigation and obstacle avoidance, we adopt the most state-of-art TAPF as our basic algorithm. The TAPF composes attractive and repulsive potential field, which attracts the USV to the goal and repulses the USV away from the obstacle. The combined potential field guides the USV to the goal while avoiding every obstacle.

*Attractive potential field.* The attractive potential and force of TAPF are given by the following equations

$$\left\|\vec{U}_{att}\right\| = \begin{cases} \frac{1}{2}K_{att}d(q,q_g)^2, & d(q,q_g) \leq d_g \\ d_g K_{att} d(q,q_g) - \frac{1}{2}K_{att}d_g^2, & d(q,q_g) > d_g \end{cases} \quad (1a)$$

$$\left\|\vec{F}_{att}\right\| = -\left\|\nabla \vec{U}_{att}\right\| = \begin{cases} K_{att}d(q,q_g), & d(q,q_g) \leq d_g \\ d_g K_{att}, & d(q,q_g) > d_g \end{cases} \quad (1b)$$

$$\left\|\vec{F}_{att}^{[x]}\right\| = \left\|\vec{F}_{att}\right\|\cos(\theta_{att}) \quad (1c)$$

$$\left\|\vec{F}_{att}^{[y]}\right\| = \left\|\vec{F}_{att}\right\|\sin(\theta_{att}) \quad (1d)$$

$$\theta_{att} = \arctan\left(\frac{y_g - y}{x_g - x}\right) \quad (1e)$$

where $\vec{U}_{att}$ and $\vec{F}_{att}$ are the attractive potential and attractive force, respectively. $K_{att}$ and $d_g$ denote two hyperparameters: coefficient of attractive and parabolic range separately. $q(x,y)$ represents the position of moving USV and $q_g(x_g, y_g)$ denotes the position of goal.

In equation (1), the attractive potential is a piecewise function, which remains parabolic within the range $d_g$ while becomes linear outside. The design of piecewise potential field aims to reduce the explosion of gradient far from the goal point because the attractive force, which is the minus gradient of potential field, is proportional to the distance to goal within $d_g$. Therefore, the attractive force outside range $d_g$ is set to be a constant $d_g K_{att}$. As the only variable in the potential field is the distance between USV and goal, the direction of attractive force $\theta_{att}$ points to the goal from the USV.

With the help of attractive field, the USV is attracted to the goal with the gradient descent method. The advantage of attractive potential field is its simplicity. The guidance only depends on real-time states rather than iteratively explore the whole map like Dijkstra or A*.

As is shown in Figure 1, the attractive potential field descends from the start point of USV to the goal position. Wherever the USV sails, the attractive force will lead the USV to the goal.

*Repulsive potential field.* The other key part of TAPF is the repulsive potential field, which keeps the USV away from obstacles such as island or floating objects. The repulsive potential and force are defined by the following equations

$$\left\|\vec{U}_{rep_i}\right\| = \begin{cases} \frac{1}{2}K_{rep}\left(\frac{1}{d(q,q_{oi})} - \frac{1}{d_o}\right)^2 d(q,q_g)^n, & d(q,q_{oi}) \leq d_o \\ 0, & d(q,q_{oi}) > d_o \end{cases} \quad (2a)$$

$$\left\|\vec{F}_{rep_i}\right\| = -\left\|\nabla \vec{U}_{rep_i}\right\| = \begin{cases} \left\|\vec{F}_{rep_i}^{[1]} + \vec{F}_{rep_i}^{[2]}\right\| & d(q,q_{oi}) \leq d_o \\ 0 & d(q,q_{oi}) > d_o \end{cases} \quad (2b)$$

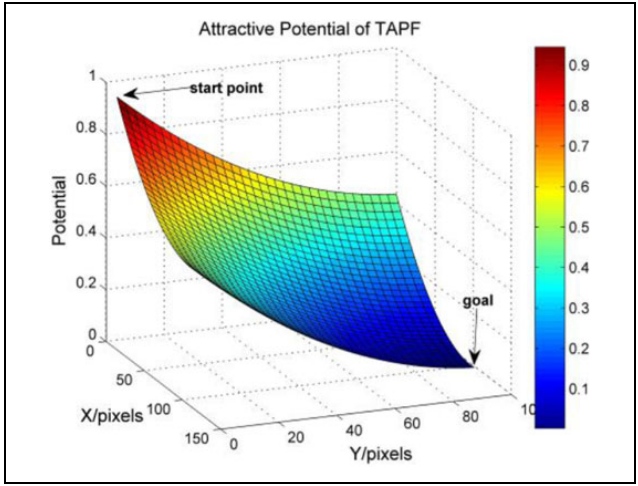

**Figure 1.** Attractive potential of TAPF. TAPF: traditional artificial potential field.

$$\left\|\vec{F}_{rep_i}^{[1]}\right\| = K_{rep}\left(\frac{1}{d(q,q_{oi})} - \frac{1}{d_o}\right)\frac{d(q,q_g)^n}{d(q,q_{oi})^2} \quad (2c)$$

$$\left\|\vec{F}_{rep_i}^{[2]}\right\| = -\frac{nK_{rep}}{2}\left(\frac{1}{d(q,q_{oi})} - \frac{1}{d_o}\right)^2 d(q,q_g)^{n-1} \quad (2d)$$

where $U_{rep_i}$ and $F_{rep_i}$ represent the repulsive potential and force of $i$th obstacle. Three hyperparameters, $K_{rep}$, $d_o$ and $n$ denote the coefficient of repulsive potential, repulsive effective range and repulsive decay of goal. Especially, $q_{oi}$ denotes the closest point on the obstacle $i$ to the USV.

We could easily find the repulsive force composes two parts $F_{rep}^{[1]}$ and $F_{rep}^{[2]}$, which represent the force from obstacle to USV $\theta_{rep}^{[1]}$ and from USV to goal $\theta_{rep}^{[2]}$. Figure 2 shows the components of repulsive force $F_{rep}^{[1]}$ and $F_{rep}^{[2]}$, while the two forces have a different function.

The force $F_{rep}^{[1]}$ is the typical repulsive force from the obstacle to hold back the USV. However, the second force $F_{rep}^{[2]}$ diminishes the attractive force to avoid non-reachable goal with the obstacle nearby. When the obstacle is much too close to the goal, where attractive force is very small, repulsive force might totally block the USV from reaching the goal. Therefore, a compensatory force is introduced. Figure 2 illustrates the resultant of repulsive field.

Consequently, the resultant repulsive force was decomposed into $x$ and $y$ directions, which are computed as the following equations

$$\left\|\vec{F}_{rep_i}^{[x]}\right\| = \left\|\vec{F}_{rep_i}^{[1]}\right\|\cos\left(\theta_{rep}^{[1]}\right) + \left\|\vec{F}_{rep_i}^{[2]}\right\|\cos\left(\theta_{rep}^{[2]}\right) \quad (3a)$$

$$\left\|\vec{F}_{rep_i}^{[y]}\right\| = \left\|\vec{F}_{rep_i}^{[1]}\right\|\sin\left(\theta_{rep}^{[1]}\right) + \left\|\vec{F}_{rep_i}^{[2]}\right\|\sin\left(\theta_{rep}^{[2]}\right) \quad (3b)$$

$$\theta_{rep}^{[1]} = \arctan\left(\frac{y_g - y}{x_g - x}\right) \quad (3c)$$

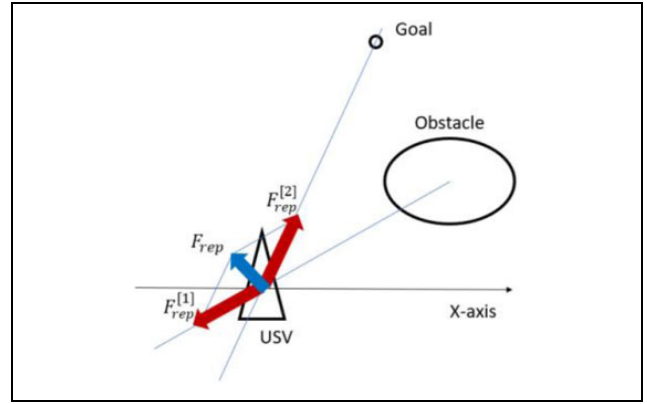

**Figure 2.** Resultant of repulsive force.

$$\theta_{rep}^{[2]} = \arctan\left(\frac{y - y_{oi}}{x - x_{oi}}\right) \quad (3d)$$

where $\vec{F}_{rep_i}^{[x]}$ and $\vec{F}_{rep_i}^{[y]}$ denote the repulsive force at $x$ and $y$ directions in the global coordinate. $q_{oi}(x_{oi}, y_{oi})$ denotes the coordinate of the closest point of $i$th obstacle.

Stated as equation (3) and Figure 3, the resultant repulsive force is the combination of every separate force.

$$\left\|\vec{F}_{rep}^{[x]}\right\| = \sum_{i=1}^{N}\left\|\vec{F}_{rep_i}^{[x]}\right\| \quad (4a)$$

$$\left\|\vec{F}_{rep}^{[y]}\right\| = \sum_{i=1}^{N}\left\|\vec{F}_{rep_i}^{[y]}\right\| \quad (4b)$$

$$\theta_{rep} = \arctan\left(\frac{\left\|\vec{F}_{rep}^{[x]}\right\|}{\left\|\vec{F}_{rep}^{[y]}\right\|}\right), \theta_{rep} \in (-\pi, \pi] \quad (4c)$$

Finally, the combined attractive and repulsive potential field generate the optimal path by gradient descent, as shown in Figure 4. This is one simple scenario that the USV avoids the obstacle island to the goal. From the start point, the USV is attracted by attractive force direct to the goal, while within the effective area of repulsive field, the USV is subjected to both attractive and repulsive force simultaneously. However, the repulsion force is stronger than attraction, so the USV must detour the repulsive field.

Admittedly, the traditional APF algorithm could lead the USV at any time, any position without iteratively adjusting the path. Nevertheless, three obvious drawbacks keep TAPF from effectively applying to real marine situation.

Firstly, vibration occurs at the joint of attraction and attraction. Subjected to more than one force, the USV has constantly repulsed away from the obstacle. However, once the USV stay away from the repulsive field, it tries to directly turn to the goal again. Therefore, the USV would easily shake its head between repulsion and attraction.

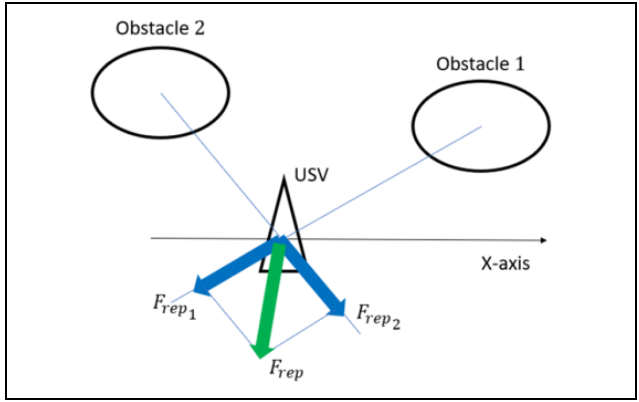

**Figure 3.** Resultant force from multiple obstacles.

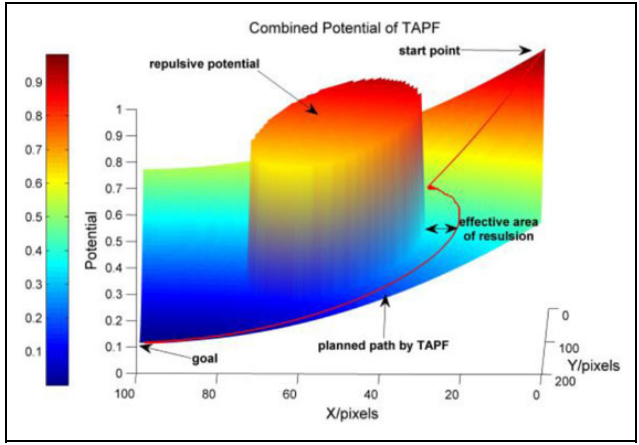

**Figure 4.** Combined potential of APF. APF: artificial potential field.

Secondly, the velocity of USV or called learning rate of gradient descent is set as a constant. Usually, the USV faces three conditions: straightway, cornering and turning around. If the USV sails on the open area without any obstacle, it should accelerate to save time, while in a complicated environment, the USV should decelerate to ensure security. However, the adaptive velocity is not considered in the TAPF.

Finally, the generated path by TAPF is not conforming with dynamics. One essential principle for path planning is smoothness or continuous derivative. But the TAPF adopts piecewise repulsive potential, which abruptly disappears outside effective area. Therefore, the USV has to turn precipitately that does not correspond to any control strategy. Otherwise, the TAPF is totally different from real USV dynamic.

To deal with the aforementioned problems, we propose three modifications and combine them into PAPF algorithm. The PAPF could both adjust velocity automatically and avoid shake or abrupt turning, ensuring a smooth, fast and secure path for high-speed USV.

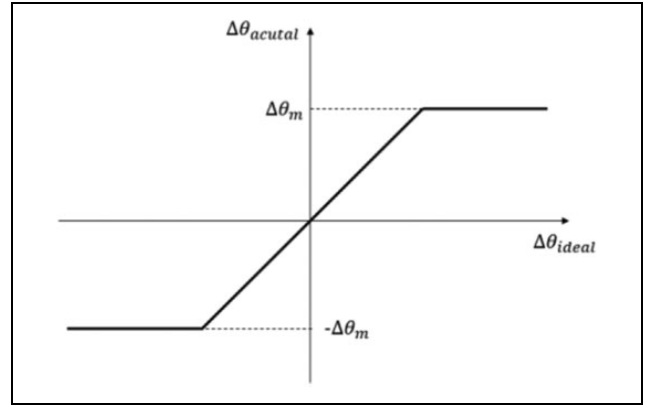

**Figure 5.** Saturation of ideal turning angle.

### Predictive artificial potential field

The PAPF composes three modifications including AL, VA and predictive potential. Finally, with the help of PAPF, the generated path would meet the requirement of USV dynamic.

*Angle limit.* In reality, the USV is not a mass point but rigid body, which has to obey rigid body dynamic. Facing an obstacle, the USV is held back by the repulsive potential to turn around immediately. However, the steering angle of USV is limited by yaw moment $M_z$ and moment of inertia $I_{zz}$. Therefore, the maximum of turning angle $\Delta\theta_m$ must obey the following dynamic equations

$$I_{zz}\dot{r} = M_z(\delta, V_v) \quad (5a)$$

$$\Delta\theta_m = r\Delta t \quad (5b)$$

where $r$ denotes the angular velocity and $\Delta t$ is the sampling period. Yaw moment $M_z$ is determined by the two variables steering angle $\delta$ and vehicle velocity $V_v$ according to the dynamic model.

If the change in ideal turning angle $\Delta\theta_{ideal}$ exceeds the AL, the actual turning angle $\Delta\theta_{actual}$ will saturate. The ideal turning angle is defined as the deviation between yaw angle at last time step $k$ and expected yaw angle at new time step $k + 1$, as shown in the following equation

$$\Delta\theta_{ideal} = \hat{\theta}_v(k+1) - \theta_v(k) \quad (6)$$

where $\hat{\theta}_v(k + 1)$ represents the current expected yaw angle, while $\theta_v(k)$ denotes the last yaw angle. Correspondingly, $\Delta\theta_{ideal}$ is the ideal turning angle of USV. Therefore, we introduced saturate adjustment of turning angle as equation (7) and Figure 5.

$$\Delta\theta_{actual} = Saturation(\Delta\theta_{ideal}, \Delta\theta_m) \quad (7a)$$

$$\theta_v(k+1) = \Delta\theta_{actual} + \theta_v(k) \quad (7b)$$

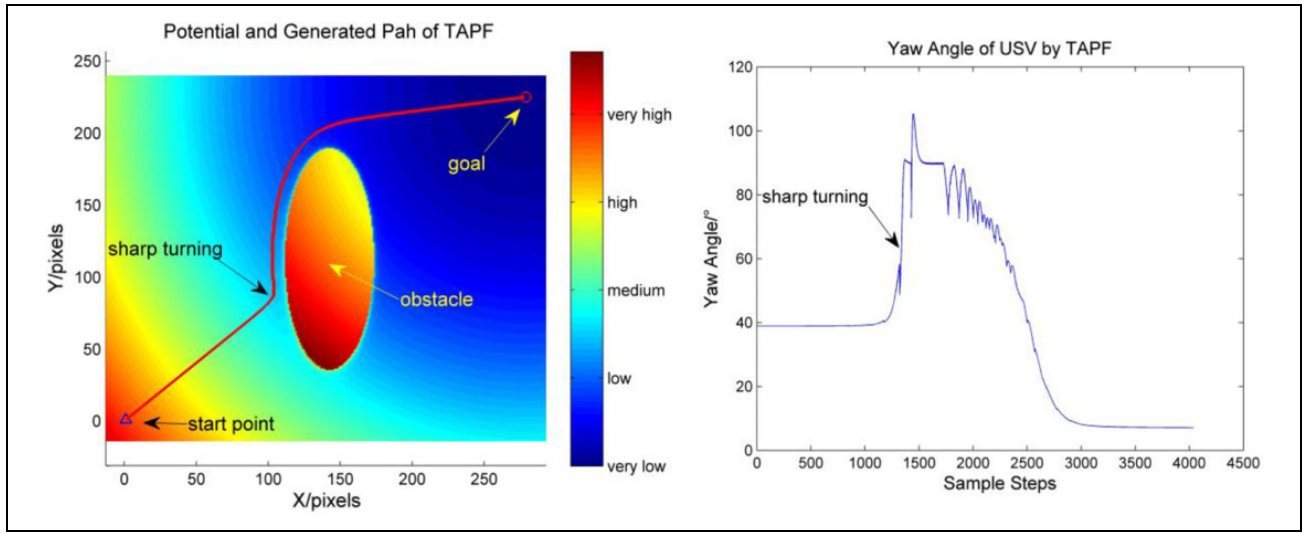

**Figure 6.** Potential and generated path of TAPF and its yaw angle profile. TAPF: traditional artificial potential field.

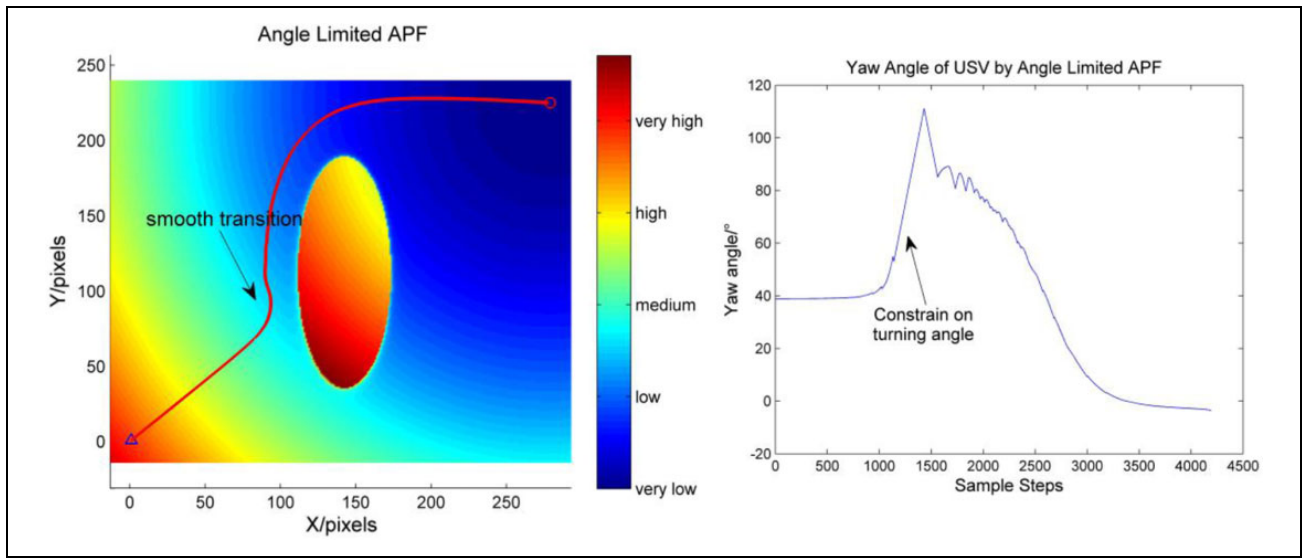

**Figure 7.** AL APF and its yaw angle profile. AL: angle limit; APF: artificial potential field.

where the saturation function is defined as in Figure 5 with output limitation of $\Delta\theta_m$.

The AL modification of APF helps to reduce vibration and abrupt turning manipulation. The sharp turning at the joint of attraction and repulsion is restricted by AL, which generates a smoother path according to the USV dynamic. An experiment shows how USV avoids the obstacle by APF with AL. The maximum turning angle is set as $20°$.

Figure 6 shows the traditional APF while Figure 7 is the angle-limited APF method. The TAPF turns sharply without any limit, which is not corresponding to dynamic model. The yaw angle graph clearly interprets the sharp turning point. Fortunately, the AL modification successfully constrains too high turning angle. However, the AL causes farther detour, which costs 5% more time than traditional APF. Therefore, we adopt another modification called VA to cut short sailing time.

*Velocity adjustment.* Basically, the control strategy cannot be totally separated from navigation because trajectory planning, which adds velocity information on the original path, is more commonly applied in the USV control system. Therefore, we introduce the VA method according to the original path. We divide the motion into three conditions: straightway acceleration, feasible cornering and turning-around deceleration.

In Figure 8, the ideal turning angle $\Delta\theta_{ideal}$ is defined the same as AL. Similarly, we compare the value of ideal turning angle with two threshold and adopt different VA. The pseudo-code of VA and AL is shown as Algorithm 1.

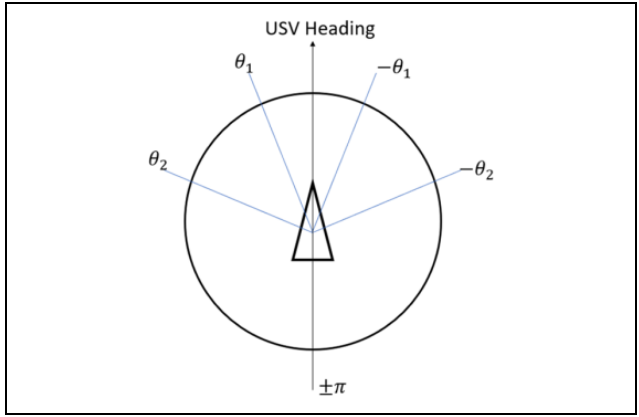

**Figure 8.** Definition of angle threshold.

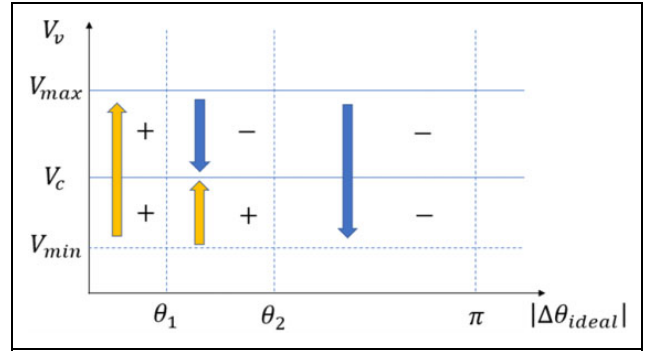

**Figure 9.** Rule of VA. VA: velocity adjustment.

**Algorithm 1.** Velocity adjustment.

---

*If* $|\Delta\theta_{ideal}| \leq \theta_1$

   *Accelerate* to $V_{max}$

*elseif* $\theta_1 < |\Delta\theta_{ideal}| \leq \theta_2$

   *Regress* to $V_c$

*elseif* $\theta_2 < |\Delta\theta_{ideal}|$

   *Decelerate* to $V_{min}$

*end*

---

where $\theta_1$ and $\theta_2$ are two thresholds that separate ideal turning angle into three parts. $V_c$ denotes the regular cursing speed of USV, while $V_{max}$ and $V_{min}$ denote the normal maximum and minimum velocity of USV. Usually, $V_{max} = 1.5 \sim 2V_c$ and $V_{min} = 0.5 \sim 0.7V_c$. Besides, the 'Regress to $V_c$' means either accelerating when speed is lower than $V_c$ or decelerating when speed is higher than $V_c$. Figure 9 more specifically illustrates how VA works.

In Figure 9, plus sign '+' denotes acceleration and minus '−' denotes deceleration at each block. The ideal turning angle might be large when facing strong repulsive field but defined within 0°–180°. In addition, the acceleration and deceleration ability depend on the dynamic model. We implement the same scenario of avoiding obstacle, as shown in Figure 10.

In Figure 10, the VA cooperates well with AL and generates a similar path. However, from the velocity profile on the right side, the velocity is adjusted automatically according to the turning angle. The path composes three main parts, two straight ways and a complex turning in the middle. Firstly, we set the cruising velocity as 0.1 pixel per step and the USV accelerates to the maximum in the straightway. Then, the USV decelerates to normal cursing speed to turn, which helps avoid directly crash into obstacle with too high speed. By experiment, we find the safe speed to avoid a collision is under 0.13 pixel per step, so the USV must slow down to 0.1 pixel per step at that position. Finally, having passed the turning area, the USV speeds up again and reach the goal.

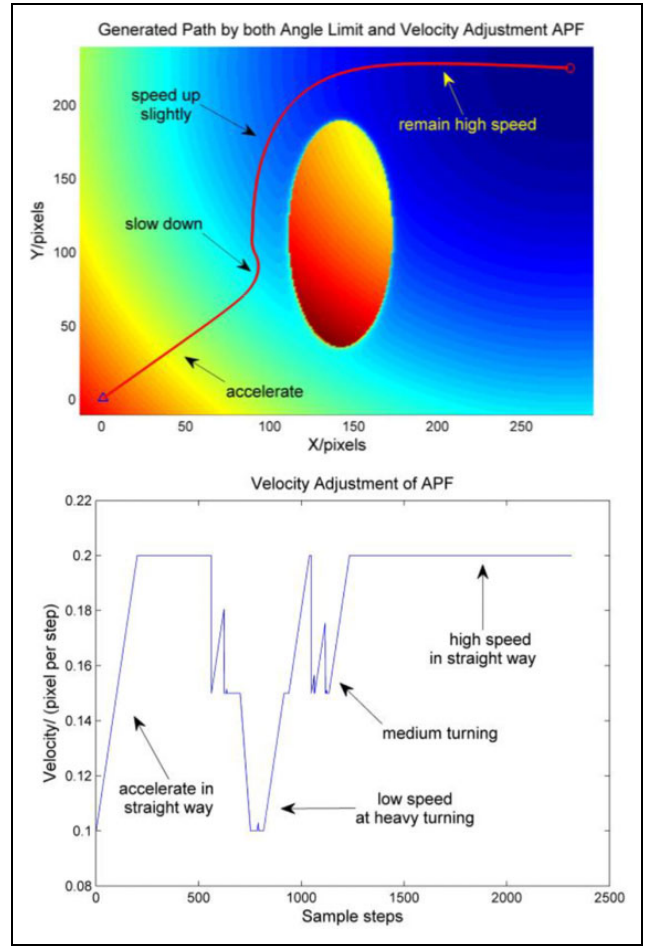

**Figure 10.** VA APF and its velocity profile. VA: velocity adjustment; APF: artificial potential field.

As the velocity before turning is higher than coursing speed, the USV needs higher turning AL to decelerate. The AL and VA methods ensure less sampling steps and maximum turning angle at the same time. The specific analysis will be together with the whole algorithm in the next section.

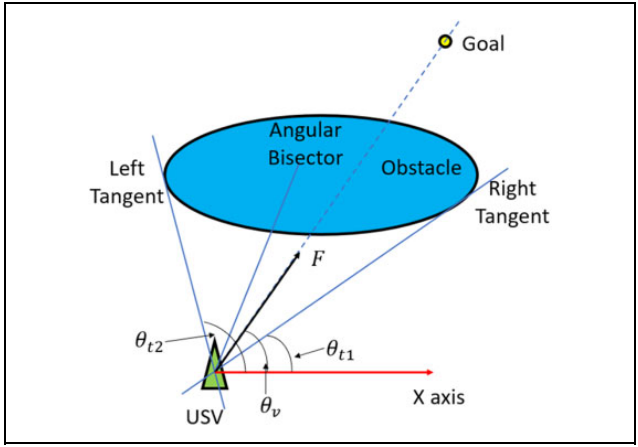

**Figure 11.** Angles defined in the predictive potential.

With the help of AL and VA, the performance of USV has been greatly improved. However, the USV still has to take a sharp turning close to the obstacle. One possible solution is to expand the repulsive potential range and intensity, while the USV must detour a large path instead. Besides, if the goal is very close to the obstacle, a strong repulsive field may cover the goal, which makes the goal unreachable. Therefore, we propose predictive potential to compensate local sharp turning.

## Predictive potential

The essential idea of predictive potential is to manipulate USV to avoid the obstacle in advance. If there is any obstacle in the way, the USV should avoid ahead of time rather than taking an abrupt turning in the range of repulsive potential. Therefore, we adopt two variables: angle deviation $|\theta_{tb} - \theta_v|$ and distance to obstacle $d(q, q_{oi})$ to construct predictive potential. Figure 11 illustrates the definition of predictive variables.

In Figure 11, $F$ denotes the combined attractive and repulsive force and $\theta_v = \theta_{ideal}$ is the ideal yaw angle of vehicle. From the position of USV, we find two tangent lines to the facing obstacle and $\theta_{t1}$ and $\theta_{t2}$ denote smaller and larger tangent angles. In addition, we define the angular bisector of two tangent lines as equation (8a) and the angle deviant between USV yaw angle $\theta_v$ and bisector angle $\theta_{tb}$ is expressed as equation (8b).

$$\theta_{tb} = \frac{\theta_{t1} + \theta_{t2}}{2} \quad (8a)$$

$$\Delta\theta_v = \theta_{tb} - \theta_v \quad (8b)$$

Subjected to distance to obstacle and angle deviant, the predictive potential can be expressed as the following equations

$$\left\|U_{prd}\right\| = K_{prd}\exp\left(-\frac{d(q, q_{oi})}{d_{prd}} - |\Delta\theta_v|\right) \quad (9a)$$

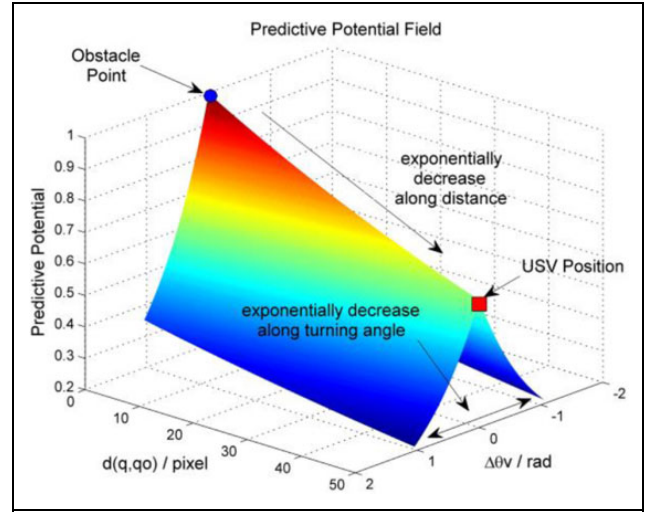

**Figure 12.** Predictive potential field.

$$\vec{F}_{prd} = -\nabla U_{prd} = \vec{F}_{prd}^{[1]} + \vec{F}_{prd}^{[2]} \quad (9b)$$

$$\left\|\vec{F}_{prd}^{[1]}\right\| = \frac{U_{prd}}{d_{prd}} \quad (9c)$$

$$\left\|\vec{F}_{prd}^{[2]}\right\| = U_{prd} \quad (9d)$$

where $U_{prd}$ and $\vec{F}_{prd}$ denote predictive potential and force. Two hyperparameters $K_{prd}$ and $d_{prd}$ represent the predictive coefficient and effective range. The predictive potential field exponentially decreases along the distance and angle deviant, without abrupt boundary like repulsive potential. The exponential design contributes a smoother edge that avoids sharp turning. In addition, the potential force composes two parts in a different direction, whose heading are defined as $\theta_{prd}^{[1]}$ and $\theta_{prd}^{[2]}$, respectively.

$$\theta_{prd}^{[1]} = \arctan\left(\frac{y - y_{oi}}{x - x_{oi}}\right) \quad (10a)$$

$$\theta_{prd}^{[2]} = \begin{cases} \theta_{tb} - \frac{\pi}{2}, & \Delta\theta_v \geq 0 \\ \theta_{tb} + \frac{\pi}{2}, & \Delta\theta_v < 0 \end{cases} \quad (10b)$$

Equation (10) shows the direction of predictive force. The definition of $\theta_{prd}^{[1]}$ is simply from the obstacle to USV. However, $\theta_{prd}^{[2]}$ – the direction of $\vec{F}_{prd}^{[2]}$ – is piecewise according to the sign of $(\theta_{tb} - \theta_v)$ as the derivative of absolute value is determined by its sign. The distribution of predictive potential is illustrated in Figure 12.

In Figure 12, the potential is not only determined by the relationship between USV and obstacle but also by the turning angle of USV. With the increasement of $d(q, q_o)$ and $\Delta\theta_v$, the $U_{prd}$ decreases exponentially to zero. Therefore, the predictive force will drive the USV away from the obstacle in advance and generates a smoother path. Figure 13 shows the advantage of predictive force.

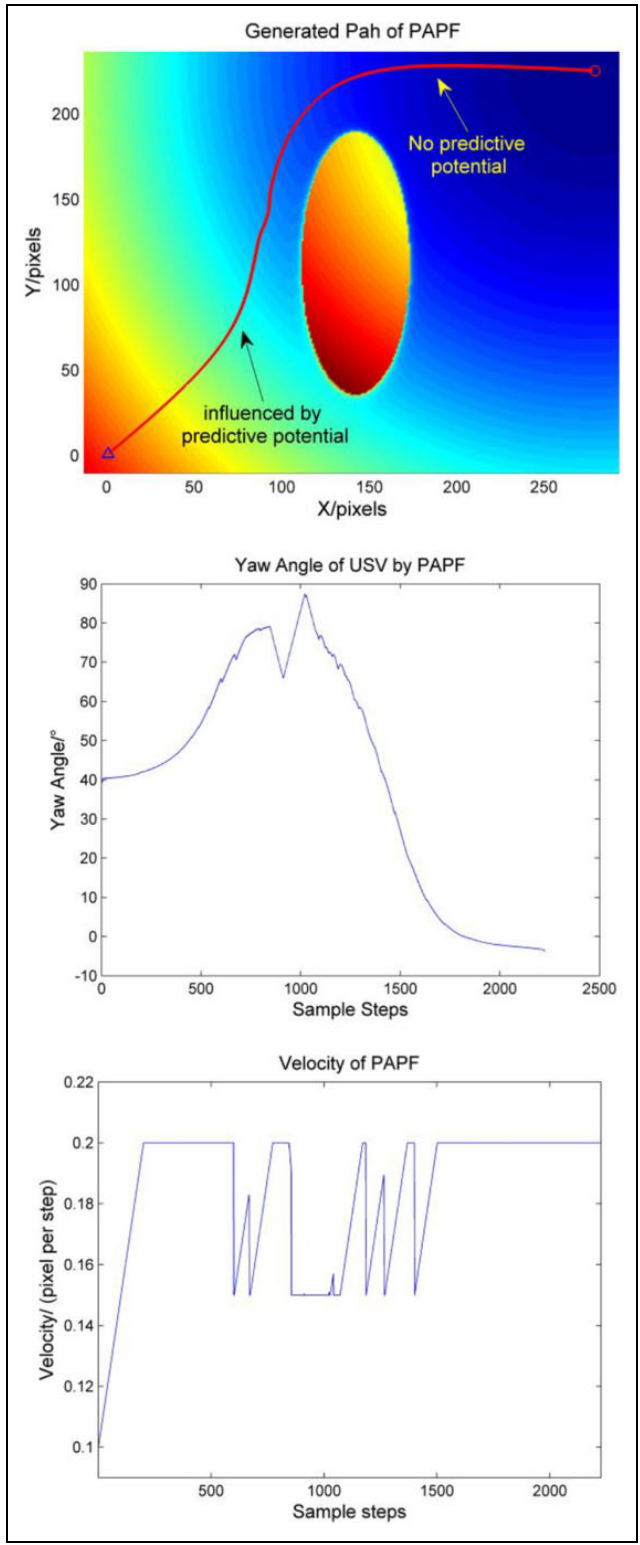

**Figure 13.** Obstacle avoidance of PAPF. PAPF: predictive artificial potential field.

Firstly, in the figure of generated path, the USV is subjected to predictive potential before repulsion. The USV is navigated to the left side smoothly, which finally results in a small turning angle. Soon afterwards, the USV passes the obstacle and directly faces the goal. Thus, there is no predictive potential anymore and the USV successfully reaches the goal.

Secondly, the yaw angle of the path becomes much smoother than traditional APF. The angle profile shows that the sharp turning once in the TAPF disappears. Even though, the USV is still under attraction and repulsion, which causes local shaking, the turning angle is constrained at a low level.

Moreover, the velocity could be much higher than TAPF as less sharp turning is required. From the velocity profile, the average velocity is up to 0.17 pixel per step, which even higher than the VA method. Therefore, predictive potential enhances the performance of the other two modification.

With the three modifications above, we combine them into PAPF. The basic potential includes attractive, repulsive and predictive potential, while AL and VA assist better path.

### Utility of predictive potential field

The whole algorithm of PAPF is expressed in Chart 1. Although we introduce three modifications from AL to predictive force, the implement consequence is inverse. The predictive potential is added up to attractive and repulsive potential and generates an ideal turning angle, followed with VA and AL. Therefore, we obtain a new turning angle and velocity to implement gradient descent update. Iteratively planning trajectory, the PAPF finally navigates the USV to the goal. A smooth and safe path is consequently generated.

To analyse the performance of PAPF, we compare the maximum turning angle and total time spent in the scenario above. As given in Table 1, it is obvious that traditional APF requires very large turning angle and costs much time. With the help of AL, the maximum turning angle shrinks to only 3% of the original value; however, the total time slightly rises. The third algorithm combines AL and VA and undoubtedly cut short 45% of time. But the higher speed leads to a bit higher turning angle. Thus, finally, the predictive APF that combines all modifications improves both turning angle and total time. The smooth path does help to reduce sharp turning, which in return ensures higher speed.

In addition to enhancing the performance of path planning, the PAPF can also solve the concave local minimum problem. The concave local minimum usually happens at the concave obstacle and the USV would be stuck into the concave part. A typical crescent-shaped obstacle would generate local minimum area. Figure 14 illustrates that TAPF cannot predict the obstacle in its way, thus finally gets trapped in local minimum. This question usually might be solved by searching the whole map or turning around in

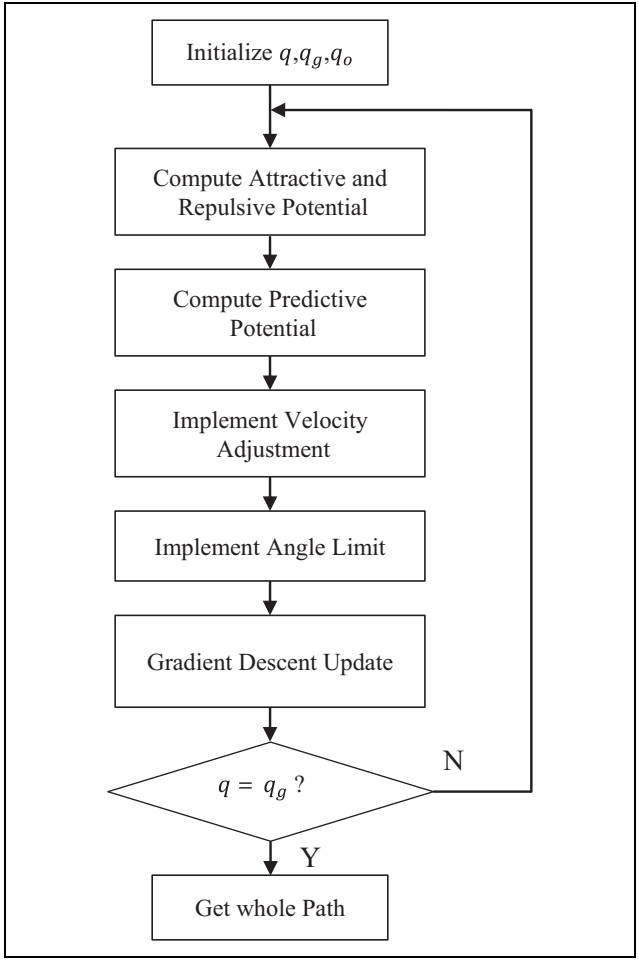

**Chart 1.** Algorithm of PAPF. PAPF: predictive artificial potential field.

**Table 1.** Maximum turning angle and total time by four methods.

|  | Maximum turning angle per step (°) | Total time (steps) |
|---|---|---|
| Traditional APF | $134.5 \times 10^{-3}$ | 4182 |
| Only AL APF | $3.5 \times 10^{-3}$ | 4191 |
| Both AL and VA APF | $5.2 \times 10^{-3}$ | 2314 |
| Predictive APF | $3.5 \times 10^{-3}$ | 2226 |

APF: artificial potential field; AL: angle limit; VA: velocity adjustment.

the local minimum. However, PAPF could easily avoid this problem.

Figure 15 explains how PAPF make the USV navigate the crescent-shaped obstacle. The effective area of predictive force is far wider than repulsive potential, which helps the USV to detect the obstacle and take action in advance. Therefore, the USV does not sail into local minimum area but detours from the right side.

In addition to the concave local minimum problem, the configuration of goal also influences the reachability of TAPF. For example, if the obstacle is not round enough, which can generate multidirectional potential, the USV might be stuck when facing an obstacle and the goal at the same time.

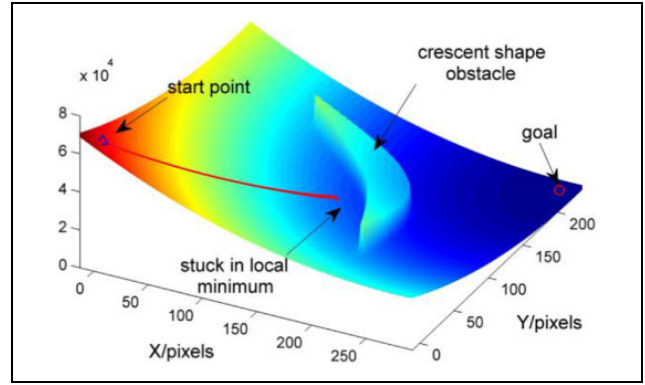

**Figure 14.** Concave obstacle avoidance of TAPF. TAPF: traditional artificial potential field.

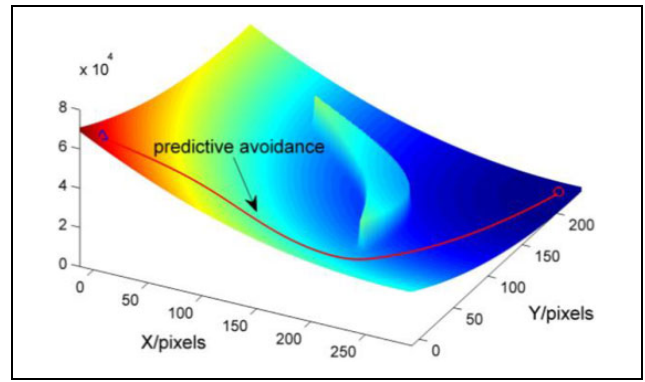

**Figure 15.** Concave obstacle avoidance of PAPF. PAPF: predictive artificial potential field.

Figure 16 vividly explains the reachability of TAPF. First, we simulate in the simply convex obstacle scenario with TAPF. The difference is we set 11 different goals from the same start point. Figure 16 shows that the upper 2 and lower 3 points are reachable, while the rest points are not reachable. The trap happens when the USV, obstacle and goal are in the same direction. Thus, the gradient generates from potential vanishes at that point.

Another question is that the USV cannot choose the better side to avoid the obstacle. The path is mostly determined by the shape of obstacle and the relative position of goal. Fortunately, PAPF could easily reach every goal with an intelligent path.

Figure 17 shows the same task with PAPF algorithm. Different from the TAPF, PAPF successfully navigates the USV to all the goals. Besides, the PAPF determines to avoid the obstacle from left or right side according to the goal position in advance. The predictive ability enables the USV to make decision not only on static

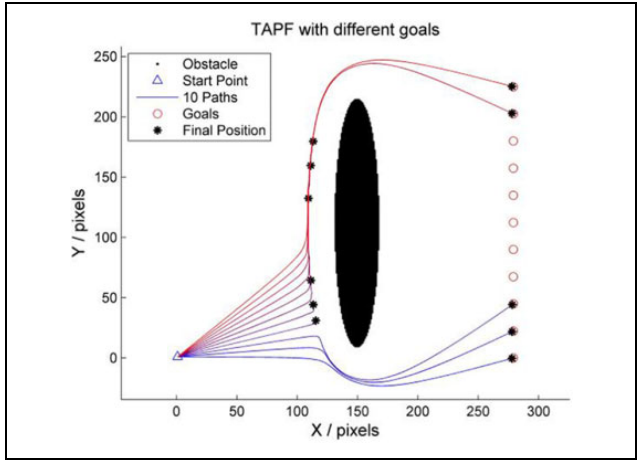

**Figure 16.** Reachability of TAPF. TAPF: traditional artificial potential field.

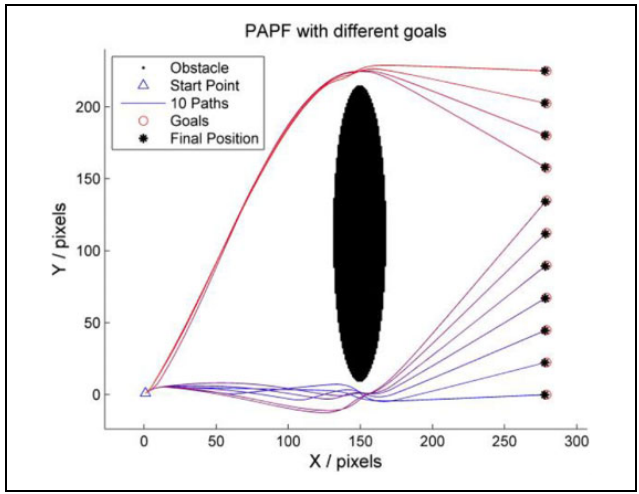

**Figure 17.** Reachability of PAPF. PAPF: predictive artificial potential field.

potential but also on the dynamic potential that considering the relative position.

In general, PAPF does well in local trap avoidance and increases reachability even with a complex scenario like concave obstacle and gradient vanishing. So far, we have introduced the basic idea and function of PAPF and we will discuss and analyse the principle and extension of PAPF.

## Discussion

### Comparison between repulsive and predictive potential

It seems that the repulsive potential is similar to the predictive potential as both of them are devised to keep the USV away from the obstacle. However, their principle and function are totally different. The repulsive potential is static and only depends on the position of obstacle. Thus, the repulsive potential is determined initially and never changes. Wherever the USV is, the repulsive potential remains constant and is only effective within a small area. The basic function of repulsive field is to protect the USV from crashing into the obstacle, but not wisely avoid them. So, it is normal to find an unrealistic and unreasonable path the repulsive potential navigates. If the shape of obstacle is not regular, consequently, the generated path will be strange as well. The reason is repulsive potential has a clear edge and disappears abruptly.

On the contrary, the predictive potential is variant according to the relative position of obstacle and the USV, which is to predict whether the future path is feasible. If the ideal direction of USV is blocked by the obstacle, the predictive potential is activated and generates lateral force to deflect direction. In this way, the USV will never reach a local minimum area and sails along a smoother path because the predictive potential is defined as an exponential function. The central area of predictive potential rises to repulse the USV and decreases smoothly to the boundary of obstacle. As long as the USV is heading the obstacle, it receives predictive force. Therefore, the predictive potential navigates in long distance and predicts the best manipulation dynamically.

### APF with time information

Whether local or global navigation, it is reasonable to consider path planning with time information, namely trajectory. The trajectory planning composes both position and time information, where angular velocity and planar velocity are two key involved variables.

The biggest problem of grid-based algorithm is the absence of time information. The moving robot on the discrete map can only move one-unit distance at each time. To enhance the accuracy, we have to sample more points on the map, which, on the contrary, raises computation exponentially. Besides, the discrete grid is not consistent with continuous control tasks. It is arbitrary to assume that the moving robot can move along the square path on the grid, especially on a very high speed.

Fortunately, APF never meets the discrete dilemma. Fortunately, APF never meets the discrete dilemma, because APF is not a whole map algorithm but local one. At each sampling point, the moving robot only considers the potential field at this place. Thus, APF is extremely computational efficient. In addition, APF is a grid-free method, which means continuous moving is preferred. With these two advantages, we proposed AL and VA algorithm to add time information.

The AL modification actually restricts the angular velocity per step, while VA controls the best velocity strategy. Thus, when we have planned the whole trajectory, we not only obtain path, but also yaw angle and velocity profile. Maybe the real dynamic model does not totally match the predictive APF, while the navigation can be updated online swiftly.

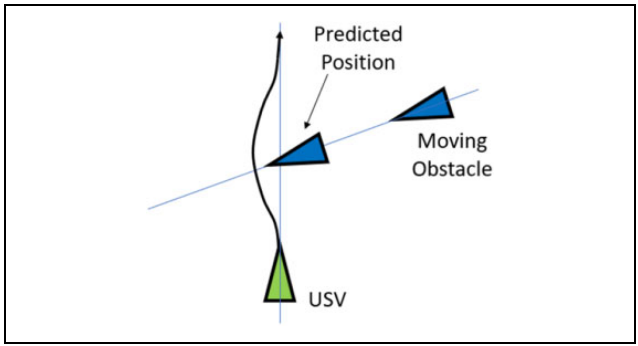

**Figure 18.** Prediction of moving obstacle.

### Moving obstacle prediction

The predictive APF performs well in static obstacle avoidance as is presented above. In fact, the PAPF can also be applied to moving obstacle avoidance. For example, the USV needs to sail across a harbour, where plenty of naval crafts sail everywhere, or the USV is under attack from a fast-moving torpedo. The predictive potential navigates the USV to avoid the possible obstacle in the future and predicts its trajectory.

Figure 18 shows the idea of moving obstacle prediction (MOP). The MOP composes two parts: to predict the USV movement and the obstacle movement. If the obstacle is going to block the USV, even if they are far away right now, the MOP algorithm generates predictive potential to avoid the future block. The specific implement of MOP is very similar to basic predictive potential, including distance and direction estimation and computation.

However, it is hard to estimate the movement of unknown obstacle for they might not move straight or be under disturbance. Hence, the MOP should adopt more information even like state estimation to avoid moving obstacle. Besides, if multiply moving obstacles exist in the same map, it is more complicated to decide how to predict and avoid all of them separately. Therefore, the MOP will be a worthwhile topic for attention.

## Conclusion

First, we review the importance of USV and the advantage of APF for path planning. Then, we propose the PAPF to navigate high-speed USV across obstacles. The PAPF adopts three modifications including AL, VA and predictive potential to generate a smoother and reasonable path. Besides, the predictive potential well predicts the obstacle in advance and choose the best strategy to avoid them. Simulation and experiments show that PAPF can effectively solve the concave local minimum problem and enhance the reachability. Finally, we analyse the principle and extension of PAPF in moving obstacle avoidance.

### Acknowledgements

The authors thank the colleagues for their constructive suggestions and research assistance throughout this study. The authors also appreciate the associate editor and the reviewers for their valuable comments and suggestions.

### Declaration of conflicting interests

The author(s) declared no potential conflicts of interest with respect to the research, authorship, and/or publication of this article.

### Funding

The author(s) disclosed receipt of the following financial support for the research, authorship, and/or publication of this article: This work was supported by the National H863 Foundation of China (Grant Numbers 11100002017115004, 111GFTQ2019115006, and 111GFTQ2018115005) and the National Natural Science Foundation of China (Grant Numbers 61473015 and 91646108).

### ORCID iD

Ce Hao  https://orcid.org/0000-0003-1780-5166